\documentclass[11pt]{amsart}
\usepackage{amsbsy,amssymb,amscd,amsfonts,latexsym,amstext,delarray,
amsmath,graphicx} 
\numberwithin{equation}{section}

\def\bF{{\mathbb F}}

\def\cL{{\mathcal L}}
\def\cM{{\mathcal M}}

\def\R{{\mathbb R}}

\title{Heat Kernel analysis of Syntactic Structures}
\author{Andrew Ortegaray, Robert C.~Berwick, and Matilde Marcolli}

\address{California Institute of Technology \\ USA}
\email{aortegar@caltech.edu}
\address{Massachusetts Institute of Technology \\ USA}
\email{berwick@csail.mit.edu}
\address{California Institute of Technology \\ USA \newline \indent
Perimeter Institute for Theoretical Physics \\ Canada \newline \indent
University of Toronto \\ Canada}
\email{matilde@caltech.edu}
\date{}

\begin{document}
\maketitle

\begin{abstract}
We consider two different data sets of syntactic parameters and we discuss how
to detect relations between parameters through a heat kernel method developed by
Belkin--Niyogi, which produces low dimensional representations of the data, 
based on Laplace eigenfunctions, that preserve neighborhood information. 
We analyze the different connectivity and clustering structures that 
arise in the two datasets, and the regions of maximal variance in the two-parameter 
space of the Belkin--Niyogi construction, which identify preferable choices of
independent variables. We compute clustering coefficients and their variance.
\end{abstract}

\section{Introduction: the Geometry of Syntactic Features}

The Chomskian Generative Linguistics approach represented the
first serious program aimed at a mathematical study of natural languages. 
The development of the mathematical theory of formal languages, for instance, 
can be seen as having partly arisen as a spinoff of this program. The
aspect of this broad viewpoint that we are more directly interested in here
is the concept of syntactic parameters, namely the idea that syntactic structures 
of natural languages can be ``coordinatized" by a set of binary variables. 
This part of the broader ``Principles and Parameters"  model of syntax in essence 
postulates that syntactic structures of natural languages can be fully
encoded in a vector of binary variables, which are usually syntactic 
parameters. The idea goes back to Chomsky's seminal work \cite{Cho}, \cite{ChoLa}, 
and has since played a crucial role in Generative Linguistics. A broad
survey of the concept of syntactic parameters is given in \cite{Bak}.

\smallskip

Among the shortcomings of the model is the fact that it has not been 
possible, so far, to identify a complete set of such syntactic parameters 
and, even though extensive lists of syntactic features are recorded
for a reasonably large number of world languages, it is unclear what
relations exist between these binary variables and whether there is
a natural choice of a set of ``independent coordinates" among them.
In other words, the question can be broadly formulated as understanding
the geometry of the space of languages, viewed at the syntactic level. 

\smallskip

It is in general very difficult for linguists to collect extensive 
data about syntactic structures for a large number of languages.
There are presently some sources of data that we have been using
for our investigation. A first source we consider is the 
 ``Syntactic Structures of the World's Languages" (SSWL) 
database \cite{SSWL}, which is freely available as an online
resource. The SSWL database has the advantage of being very extensive (presently,
it includes 116 parameters and a set of 253 world languages). However,
there are issues with these data that need to be taken into account
carefully. One problem is linguistic in nature, namely the fact that some of
the choices of binary variable recorded in the SSWL database do not 
reflect what linguists typically consider to be the ``true" syntactic parameters, 
due to conflations of deep and surface structure.  The other issue 
stems from the fact that the parameters in the SSWL database are
very non-uniformly mapped across the languages recorded of the database,
with some languages 100\% mapped with all 116 parameters
and others for which only very few of the parameters are recorded. 
Thus, our data analysis has to take into consideration how to 
handle the incomplete data. We discuss this issue in \S \ref{FilterSec}.
A second recent source of data is given
by a list of 83 syntactic parameters for a set of 62 languages (mostly 
Indo-European) collected by Giuseppe Longobardi, \cite{Longo2}, 
which extends the previously available list of Longobardi and
Guardiano \cite{Longo1}. This second list of parameters has several
advantages with respect to the SSWL data: the syntactic features 
considered can be regarded as genuine syntactic parameters;
the data are much more uniformly mapped across the set of
languages considered, even though some lacunae in the data
are still present; some relations between parameters are taken
into consideration in the data. The type of relations considered
in the Longobardi data are certain forms of entailment according to which some
parameters in a language may become undefined by effect of the value of 
other parameters. This type of relation is recorded in the data
using ternary instead of binary values, with $\pm 1$ values
corresponding to the usual binary on/off values of a given
parameter, and an additional value $0$ to denote the case where
a parameter is undefined by effect of the values of one or more
of the other parameters. In first approximation, we treat the syntactic 
features recorded in the data of \cite{Longo2} as an independent
set of data with respect to the features recorded in the SSWL
database \cite{SSWL}. 

\smallskip

We will use the term ``parameters" in this paper, for simplicity, to denote quite broadly 
various sets of binary (or ternary, if an ``undefined" value is included) variables encoding 
syntactic features, both in the case of
the SSWL data \cite{SSWL} and in the case of the Longobardi data \cite{Longo2}. 

\smallskip

A natural approach, in order to investigate relations between
syntactic parameters at the computational level, is to apply
dimensional reduction algorithms to the data and identify
possible connectivity and clustering structures. In this paper
we focus on a technique developed in \cite{BeNi}, \cite{BeNi2},
\cite{BeNi3} based on the differential geometry of Laplacians and
heat kernels. 

\section{Heat Kernel and dimensional reduction}\label{HeatSec}

We recall briefly the dimensional reduction technique developed
in \cite{BeNi}, \cite{BeNi2}, \cite{BeNi3}. 
The problem addressed by this approach is generating efficient 
low dimensional representations of data sampled from a probability distribution
on a manifold. What one aims for is a method that is generally more
efficient at identifying connectivity structures in the data than typical
dimensional reduction methods like Principal Component Analysis. 
The main idea is to build a graph associated to the data points that encodes 
neighborhood information, and use the Laplacian of the graph to 
obtain low dimensional representations that maintain the local 
neighborhood information, constructed using the eigenfunctions 
and eigenvalues of the Laplacian.

\smallskip
\subsection{The Belkin--Niyogi algorithm}\label{BNalgSec}

The general setting is the following. Consider a collection of data points $p_1,\ldots, p_k$
which lie on a manifold $\cM$ embedded in a Euclidean space $\R^\ell$. One wants to find 
a set of points $y_1,\ldots, y_k$ in a significantly lower dimensional Euclidean space
$\R^m$ (with $m<< \ell$) that suitabely {\em represent} the data points $p_i$, in the sense
that relevant proximity relations are preserved.
We assume that the data points $p_i$ are binary vectors of syntactic parameters
embedded in $\R^\ell$. Thus, in particular, we can consider the Hamming distance 
$d_H(p_i,p_j)$ between data points.

\smallskip

The first step is the construction of an {\em adjacency graph}, with vertices given
by the data points $p_i$ in the ambient space $\R^\ell$. There are several possible
methods for assigning edges in the adjacency graph:
\begin{enumerate}
\item {\em $\epsilon$-neighborhood}: an edge $e_{ij}$ is assigned between the data points
$p_i$ and $p_j$ iff the distance satisfies
$$ d_H(p_i,p_j)< \epsilon , $$
\item {\em $n$-nearest neighborhood connectivity}: an egde $e_{ij}$ is assigned between
$p_i$ and $p_j$ iff $p_i$ is among the $n$ nearest neighbors of $p_j$ or viceversa,
\item {\em farthest distance connectivity}: a node $p_i$ is connected to the $n$ farthest
nodes. 
\end{enumerate}

The third method has less immediately obvious physical interpretation, but it can be used 
to isolate highly independent syntactic parameters.
 
\smallskip

Once an adjacency graph is assigned by one of the methods listed above to the
data set, the next step in the Belkin--Niyogi algorithm consists of assigning {\em weights}
to the edges of the adjacency graph. The weights used in \cite{BeNi}, \cite{BeNi2}, \cite{BeNi3}
are based on a heat kernel
\begin{equation}\label{Wij}
 W_{ij} = \exp \left(- \frac{\| x_i - x_j \|^2}{t} \right) 
\end{equation} 
assigned to an edge $e_{ij}$, with $W_{ij}=0$ if no edge is present between
$p_i$ and $p_j$. The weights depend on a heat kernel parameter $t>0$. 

\smallskip

The Laplacian of the graph is defined as the matrix $L=D-W$, where $W=(W_{ij})$
is the $k\times k$ matrix of weights and $D$ is the diagonal matrix with diagonal
entries $D_{ii}=\sum_j W_{ji}$. One considers the eigenvalue problem
\begin{equation}\label{Leigen}
 L \psi = \lambda D \psi 
\end{equation} 
where the eigenvalues are listed in increasing order, 
$0=\lambda_0\leq \lambda_1 \leq\cdots\leq \lambda_{k-1}$ and 
$\psi_j$ are the corresponding eigenvectors, viewed as fuctions
$$ \psi_i : \{1,\ldots,k \} \to \R $$
defined on set of vertices of the graph. One assumes here that the
graph is connected, otherwise the same procedure is performed
on each connected component. 

\smallskip

The following step then consists of mapping the data set via these
Laplace eigenfuctions,
\begin{equation}\label{mapLaplace}
 \R^\ell \supset \cM \ni x_i \mapsto (\psi_1(i), \ldots, \psi_m(i)) \in \R^m .
\end{equation} 
A mapping of the data set to a lower dimensional $\R^m$ is obtained
in this way by using the first $m$ eigenfunctions. That is, the
first $m$ eigenvectors $\psi_j$, ordered by increasing associated eigvenvalues, 
form the columns of a $k \times m$ transformation matrix $T$ that transforms 
the parameter vectors $p=(p_i) \in \mathbb{R}^\ell$ to dimensionally reduced 
vectors 
\begin{equation}\label{Tmap}
p'=(p'_j) \in\mathbb{R}^m  \ \ \ \  \text{ with } \ \ \ p'_j = \sum_i T_{ji} p_i.
\end{equation}
Belkin and Niyogi discussed in \cite{BeNi3} the {\em optimality} of 
these embeddings by Laplace eigenfunctions. 

\subsubsection{Dealing with incomplete syntactic parameter data}\label{FilterSec}

To account for the incomplete mapping of languages in both the SSWL database and, 
to a lesser extent, in the Longobardi data, the data sets used have been filtered.
For the SSWL data we considered only those languages for which at least $55\%$
of the parameters are mapped, and for the Longobardi data we only used the
languages that are completely recorded ($100\%$ of the parameters mapped).
For the graphical methods, we have replaced the remaining missing values in
the SSWL data with $0.5$ values, so as not to assume a specific state of
the parameter, and so that the frequency of parameter expression is not changed.

\section{Connectivity structures of syntactic features}

For each data set over $300$ graphs were generated and analyzed.
These graphs were used for the clustering and connectivity analysis 
that we discuss in the next section, to determine structure in the parameter space.

\smallskip

We discuss here a sample of the graphs obtained with the $\epsilon$-neighborhood
method and with the $n$-nearest neighborhood method and the possible linguistic 
implications of the clusters and cliques structures detected. 

\smallskip
\subsection{Relatedness structures in Longobardi's syntactic data}

\begin{figure}[h!]
\begin{center}
\includegraphics[scale=0.85]{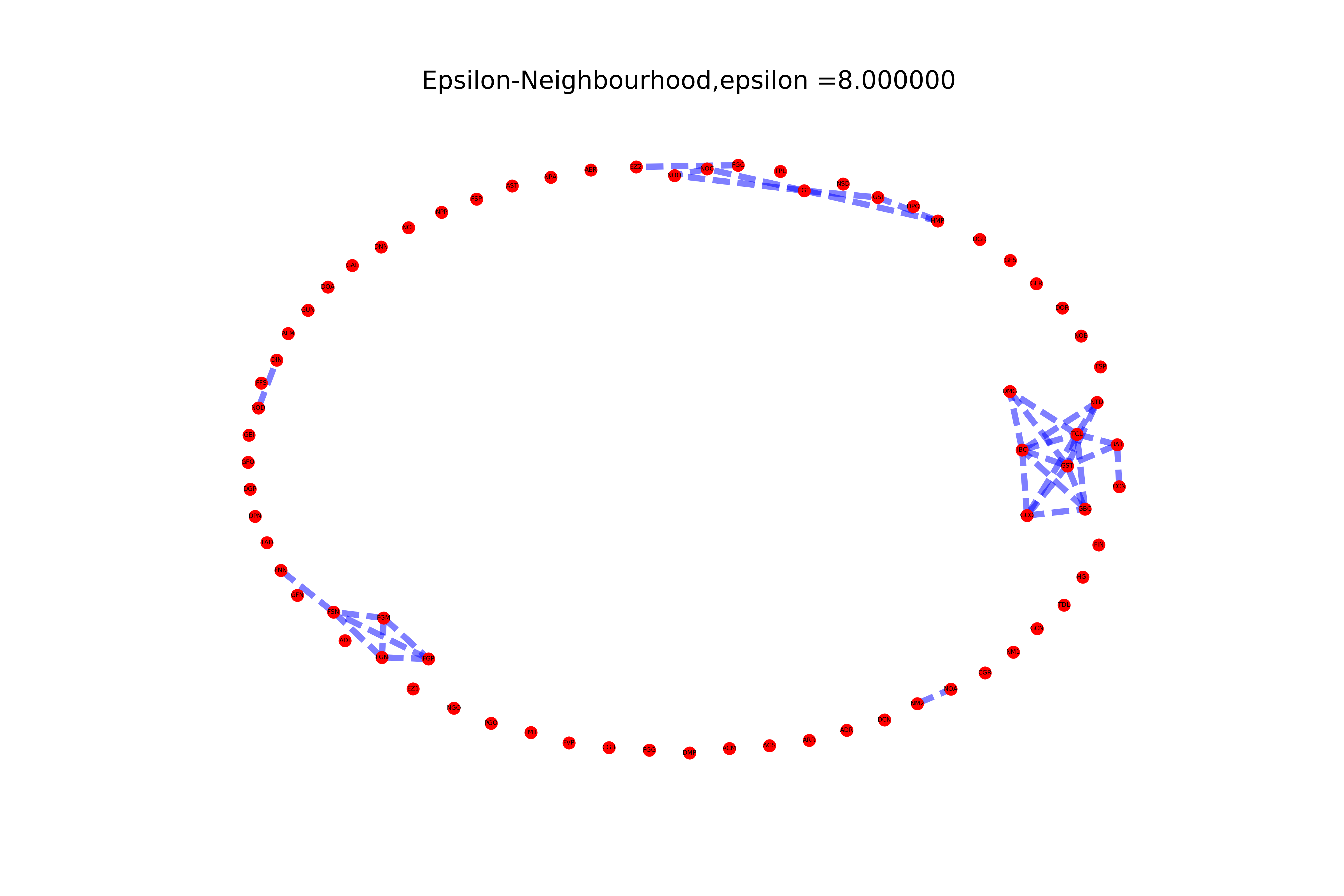}
\caption{$\epsilon$-neighborhood graph for the Longobardi dataset with $\epsilon=8$. \label{GraphsLFig}}
\end{center}
\end{figure}

\begin{figure}[h!]
\begin{center}
\includegraphics[scale=0.85]{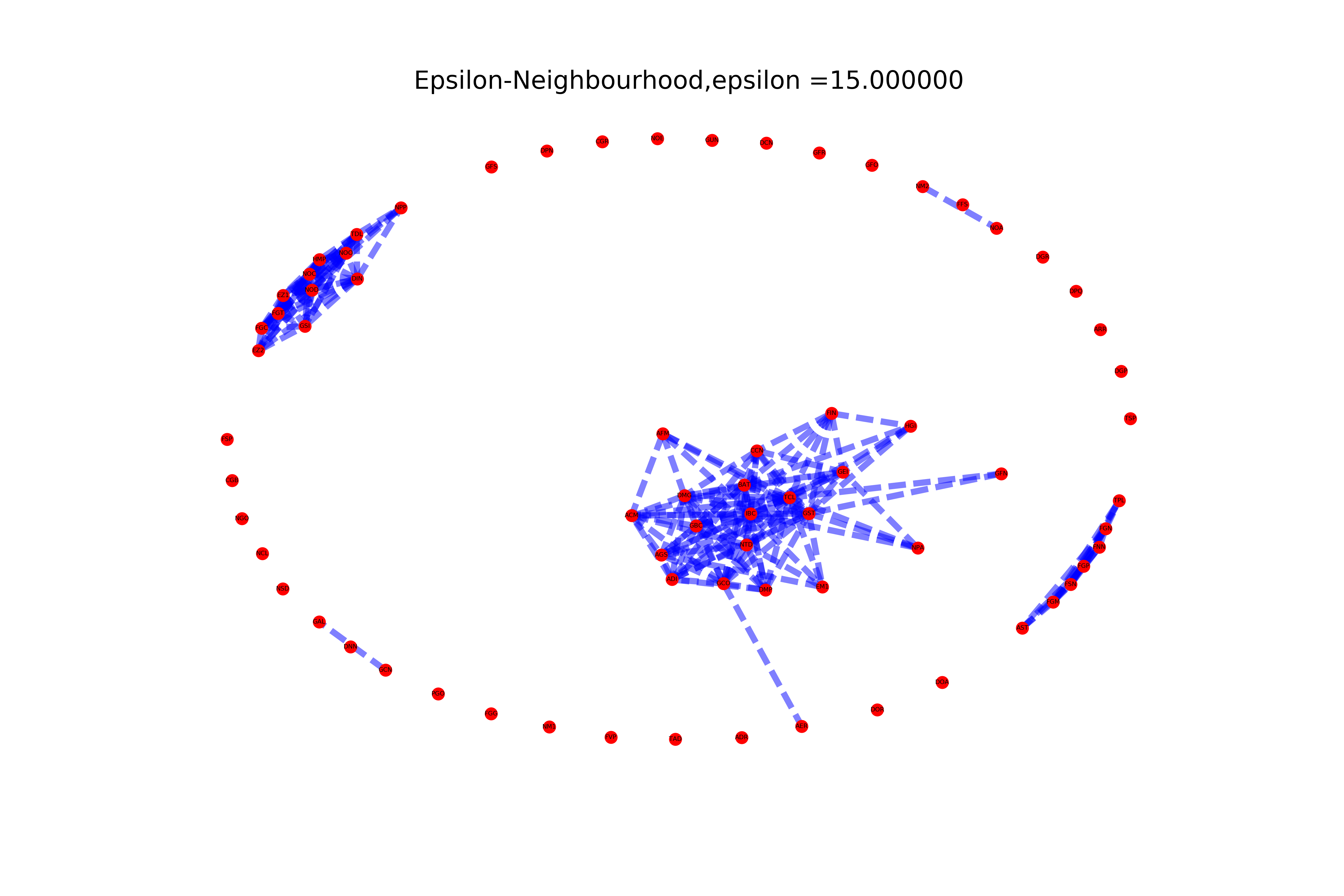}
\caption{$\epsilon$-neighborhood graphs for the Longobardi dataset with $\epsilon=15$. \label{GraphsLFig2}}
\end{center}
\end{figure}

\begin{figure}[h!]
\begin{center}
\includegraphics[scale=0.85]{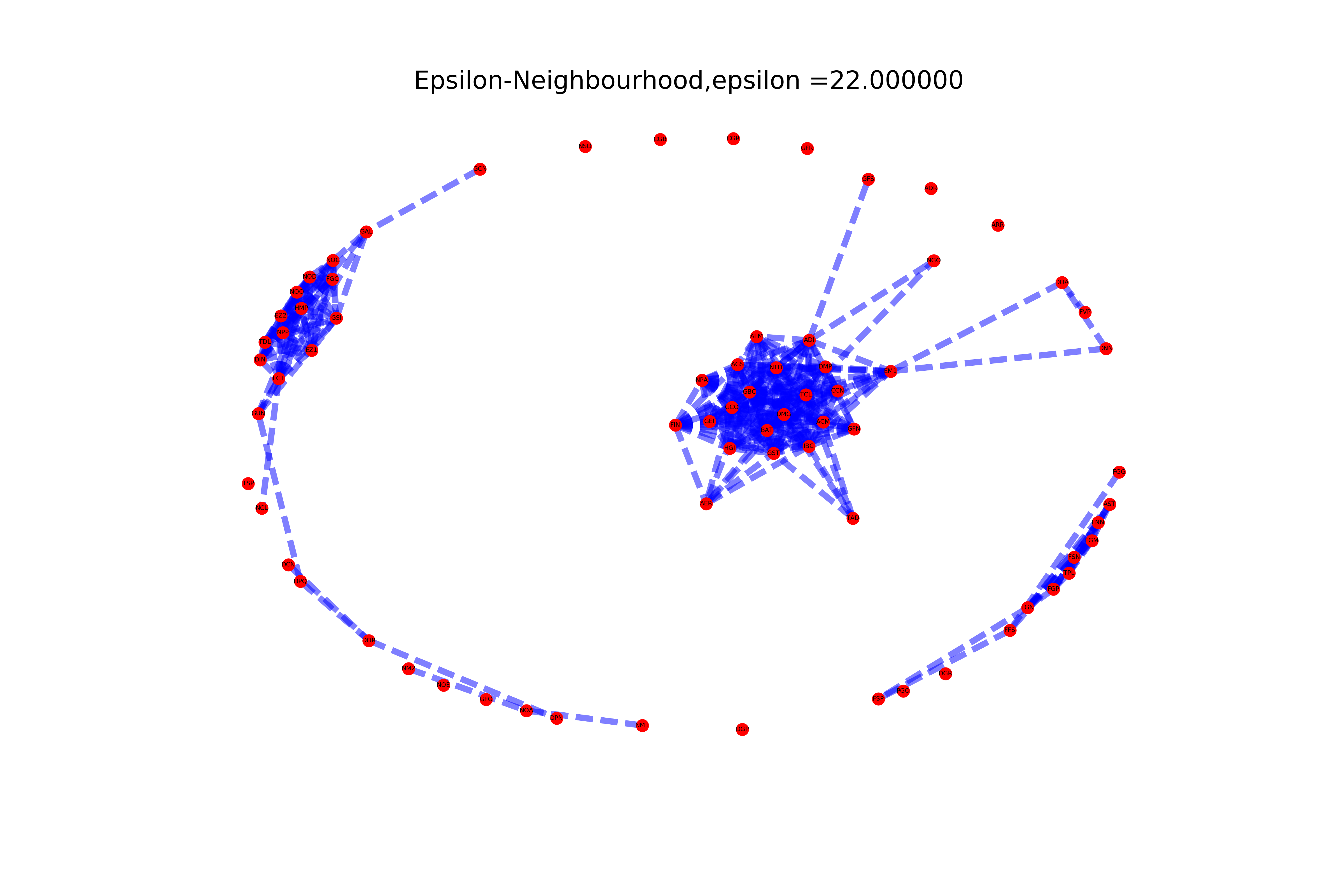}
\caption{$\epsilon$-neighborhood graphs for the Longobardi dataset with $\epsilon=22$. \label{GraphsLFig3}}
\end{center}
\end{figure}

We first consider the Longobardi dataset. We construct graphs with the
$\epsilon$-neighborhood method for different values of $\epsilon$. The
cases shown in Figures~\ref{GraphsLFig}, \ref{GraphsLFig2}, \ref{GraphsLFig3}
show the resulting graphs for $\epsilon=8$, $\epsilon=15$, and $\epsilon=22$,
respectively.

\begin{figure}[h!]
\begin{center}
\includegraphics[scale=0.25]{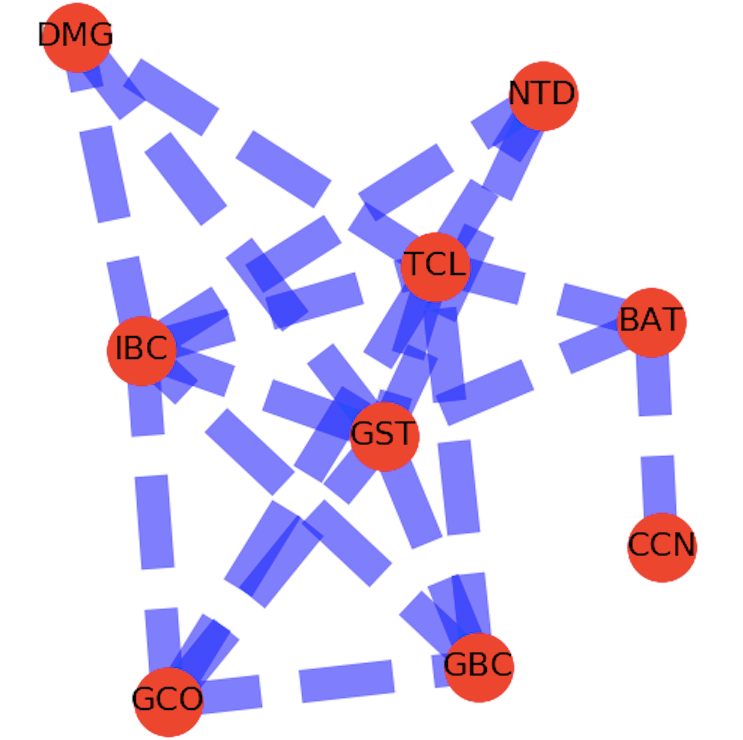}
\caption{Largest component of $G(\epsilon=8)$ for the Longobardi dataset. \label{Str1Fig}}
\end{center}
\end{figure}

In this graph one sees five relatedness structures. The two largest structures involve, respectively, $9$ and
$7$ vertices, while three smaller relatedness structures involve $5$ vertices and two sets of $2$ vertices.
The largest structure consists of the graph shown in Figure~\ref{Str1Fig}. The syntactic parameters 
related by this structure are those listed as DMG (def.~matching genitives), GCO (gramm.~collective number), GST (grammaticalised Genitive), along with other parameters: BAT, CCN, GBC,  IBC, NTD, TCL 
(see \cite{Longo2} and \cite{Longo3}).

\begin{figure}[h!]
\begin{center}
\includegraphics[scale=0.45]{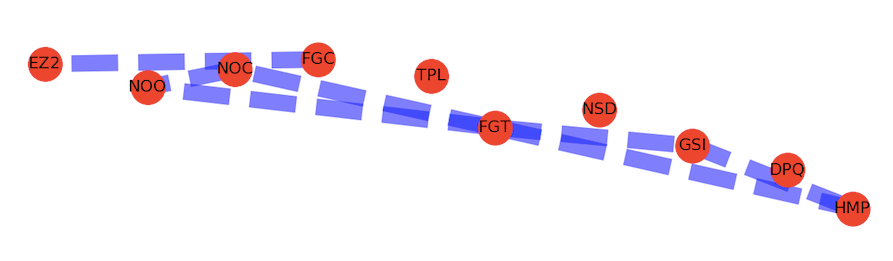}
\caption{Second component of $G(\epsilon=8)$ for the Longobardi dataset. \label{Str2Fig}}
\end{center}
\end{figure}

The second largest structure in the graph of Figure~\ref{GraphsLFig} is the graph shown in
Figure~\ref{Str2Fig}, which involves the syntactic parameters labelled EZ2 (non-clausal linker), 
FGC (gramm.~classifier), FGT (gramm.~temporality), 
GSI (grammaticalised inalienability), HMP (NP-heading modifier),  along with other parameters:
NOC, NOO.  Note that these two structures appear quite different. If we use the vertex degree (valence) as
a simple measure of centrality in a network, then we see that in the graph of Figure~\ref{Str1Fig}
the parameters TCL, GST, and IBC have valence $6$, GBC and GCO have valence $4$,
and BAT, DMG, and NTD have valence $3$, while only CCN has valence one. Thus, notes
in this network tend to have a higher degree of centrality than in the 
graph of Figure~\ref{Str2Fig}, where only the FGT and the NOC parameters have 
valence $4$, while all the other vertices have valence either one or two.  This signals
a higher degree of interconnectedness between the first group of syntactic
parameters than within the second. 

\begin{figure}[h!]
\begin{center}
\includegraphics[scale=0.45]{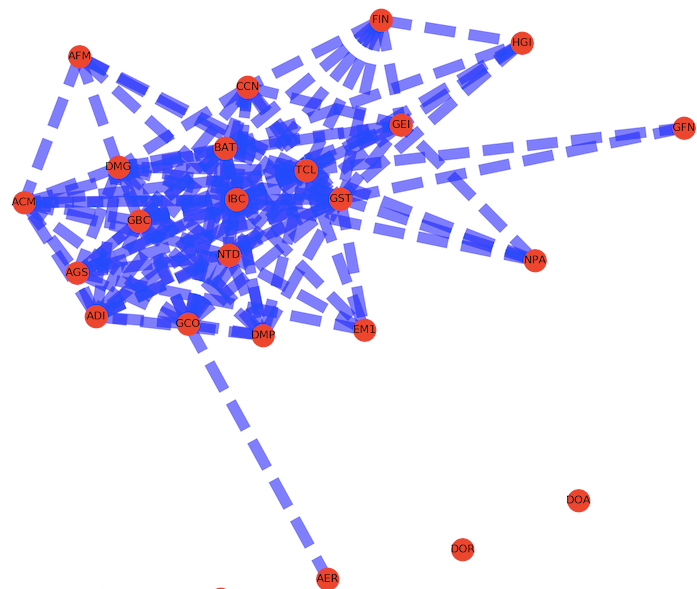}
\caption{Largest component of $G(\epsilon=15)$ for the Longobardi dataset. \label{Str3Fig}}
\end{center}
\end{figure}

When we increase the $\epsilon$ variable to $15$, we see larger relatedness structures. In particular,
we find two interesting networks. The component of Figure~\ref{Str1Fig} has grown into a much
larger component, shown in Figure~\ref{Str3Fig}, which in addition to the previous vertices 
BAT, CCN, DMG, GBC, GCO, GST,  IBC, NTD, TCL, now includes also ACM, ADI, AER,
AFM, AGS, DMP, FIN, HGI, GEI, GFN, NPA. Again, as in the previous case, most of the vertices
in this networks have high centrality and only few of them (AER, GFN, NPA) have lower degrees.
Those vertices like CCN that were peripheral for the lower value of $\epsilon=8$ in Figure~\ref{Str1Fig}
have acquired greater centrality (higher valence) at the scale $\epsilon=15$ in Figure~\ref{Str3Fig}. 
The second largest component involves the nodes DIN, EZ1, EZ2, FGC, FGT, GSI, HMP, NOC, 
NOD, NOO, NPP, TDL, and includes the network of Figure~\ref{Str2Fig} but where the
previous vertices have acquired higher centrality, 
A third smaller component appears involving connections between the parameters AST (structured APs), 
FGM (gramm.~Case), FGN (gramm.~number), 
FGP (gramm.~person), FNN (number on N), FSN (feature spread to N), TPL. 

When we further increase the neighborhood size variable to $\epsilon=22$, we see that the three
main relatedness structures identified above grow in size while still remaining three separate
components, Figure~\ref{GraphsLFig3}. These three structures also now clearly differ significantly
as network structures in terms of the centrality of nodes. The first component, which grows out of the
structure of Figures~\ref{Str1Fig} and \ref{Str3Fig} involves the nodes ACM,  ADI, AER, AFM, AGS, BAT,
CCN,  DMG, DMP,  DNN, DOA, EM1,  FIN,  FVP, GBC, GCO, GEI,  GFN, GFS, GST,  HGI,  IBC,  NGO, 
NPA, NTD,  TAD, TCL.
In this component again most of the nodes have a high degree of centrality, with only
very few peripheral nodes, such as GFS (valence $1$), NGO, DOA, DNN, FVP (valence $2$),  
TAD (valence $3$). 

\smallskip

The second component, which has grown from the component of Figure~\ref{Str2Fig}.
Unlike the previous one, this contains a subgraph consisting of nodes with low valence, 
DCN, DOR, DPN, DPQ, GCN, GFO,  NM1, NM2, NOA, NOE,  connected
through the nodes GUN and GAL to a cluster of high valence, highly interconnected
nodes, 
DIN, EZ1, EZ2, FGC, FGT, GAL, GSI, HMP, NOC, NOD, NOO, NPP, TDL, which
contains the original part of the network that already coalesced for smaller
values of $\epsilon$. 

\smallskip

While the third component grows out of the third component discussed above
for the $\epsilon=15$ graph. It involves the nodes AST, FFS, FGG, FGM, FGN, 
FGP, FNN,  FSN, FSP, PGO, TPL.    
This component also shows typical nodes of lower degrees and a lower interconnectivity.

\smallskip
\subsection{Relatedness structures in the SSWL syntactic data}

\begin{figure}[h!]
\begin{center}
\includegraphics[scale=1]{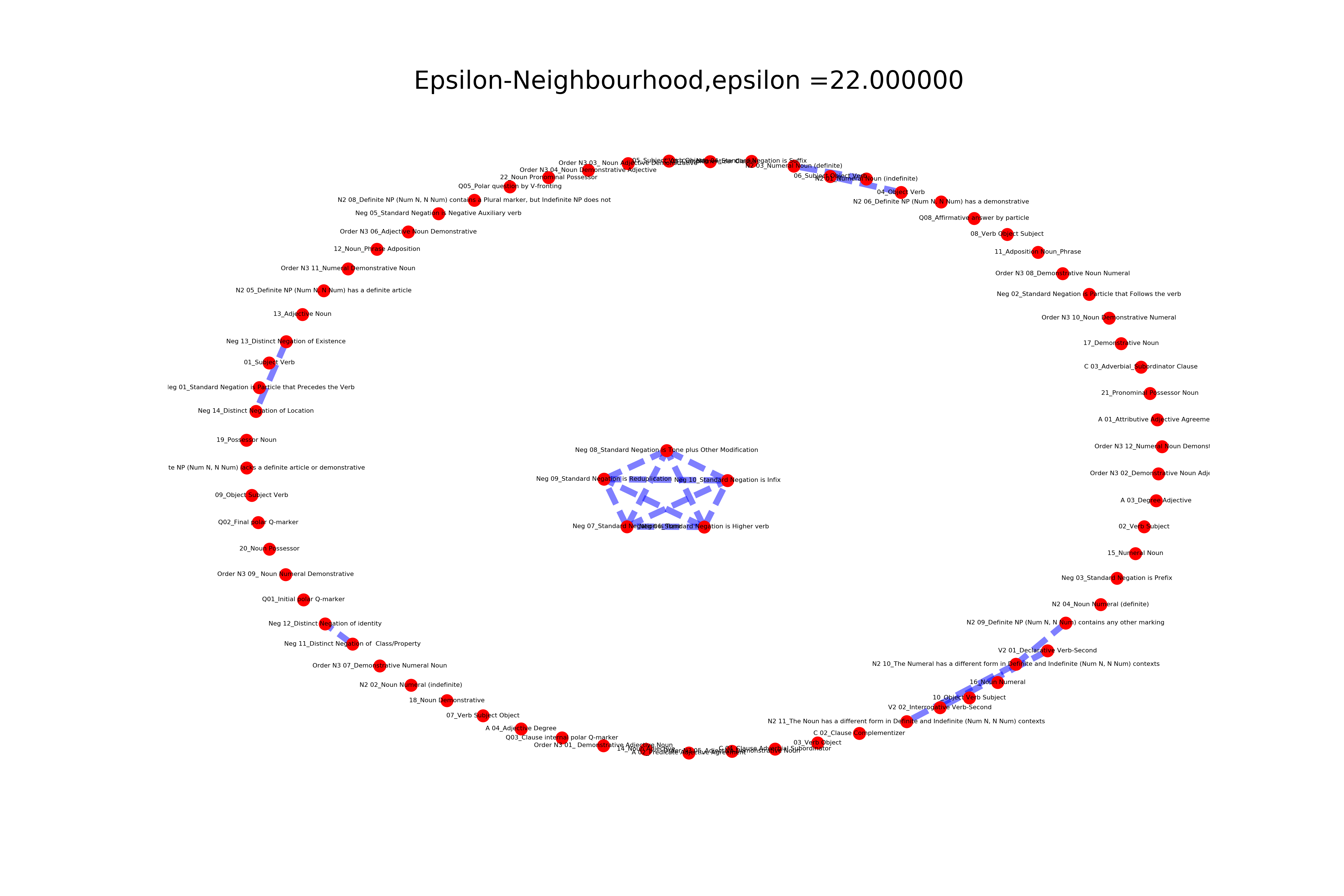}
\caption{$\epsilon$-neighborhood graphs for the SSWL dataset with $\epsilon=22$. \label{GraphSSWLFig}}
\end{center}
\end{figure}

As we discuss more in detail in \S \ref{ConnClustSec} below, connectivity and clustering 
structures in the SSWL data emerge much more slowly as a function of the neighborhood
size $\epsilon$ than in the Longobardi dataset. We can see this for instance in Figure~\ref{GraphSSWLFig},
which shows the structure of the SSWL data for the same value $\epsilon =22$ that 
we used in Figure~\ref{GraphsLFig3} for
the Longobardi data.

\begin{figure}[h!]
\begin{center}
\includegraphics[scale=0.35]{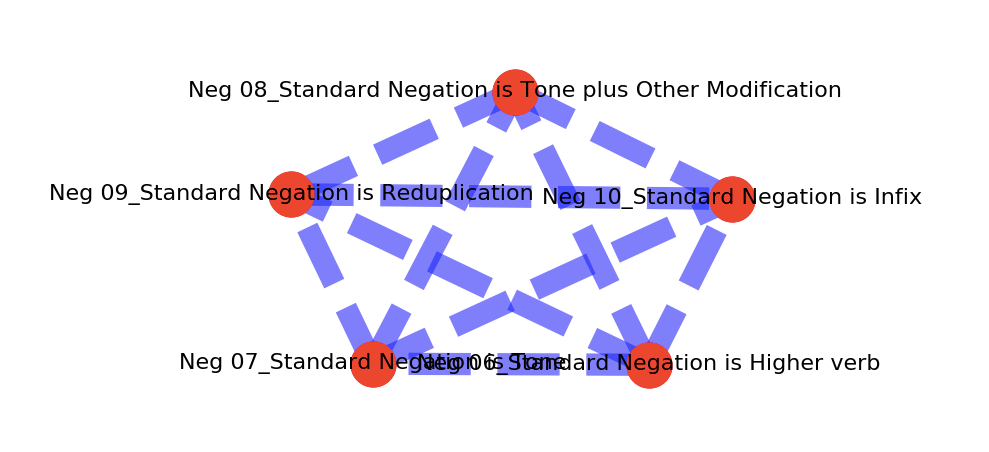}
\caption{The Standard Negation $\epsilon$-neighborhood graph for $\epsilon=22$. \label{Str1SSWLFig}}
\end{center}
\end{figure}

There are only a few small components visible at this scale in the SSWL data. One component (see
Figure~\ref{Str1SSWLFig} is a complete graph on the nodes given by the syntactic features
Neg06, Neg07, Neg08, Neg09, Neg10. These features are part of a set of binary variables (Neg01 to Neg10)
that describe properties of Standard Negation. It seems interesting that connections between the
Neg06 to Neg10 subset emerge earlier (in terms of $\epsilon$-size) than connections with the rest
of the parameters in this set. Indeed one can see by computing the graphs at $\epsilon=15$ that
the subset Neg07, Neg08, Neg09, Neg10 coalesce into a complete graph on four vertices already
at this $\epsilon$-size, while Neg06 becomes connected to this component at a larger size. This
subset of the Standard Negation parameters is indeed somewhat different in nature from the
Neg01 to Neg05 subset. The first five Standard Negation parameters describe the position of a
standard negation particle with respect to the verb (Neg01 and Neg02), whether standard negation
is expressed by a prefix or a suffix (Neg03 and Neg04) or through a negative auxiliary verb (Neg05).
The remaining set of Standard Negation parameters, which constitute the graph component of
Figure~\ref{Str1SSWLFig}, instead describe the expression of standard negation through 
predicate with a subordinate clause complement (Neg06, expressed in Polynesian languages
like Tongan), or through tone (Neg07, expressed in Niger-Congo languages like Nupe and Gu\'ebie,
or in Oto-Manguean languages like Triqui), or tone together with additional modifications to
verb form and other constituents in the negated sentence (Neg08, expressed in Niger-Congo 
languages like Basaa, Igala), by a reduplicated verb form (Neg09, expressed in Niger-Congo languages
like Eleme), or by an infix (Neg10, possibly expressed in the Muskogean language Chickasaw).
The occurrence of the graph of Figure~\ref{Str1SSWLFig} appears to indicate that these 
modes of Standard Negation more strongly correlate to one another than the other modes
described by the Neg01 to Neg05 variables.  

\smallskip

At the scale $\epsilon=22$ there is also a three vertex component involving N2-09, N2-10, N2-11, 
with one edge between N2-11 and N2-10 and one between N2-10 and N2-09, as well as several 
components consisting of two vertices joined by an edge, such as V2-01 and V2-02,
N2-01 and N2-03,  04 and 06, Neg13 and Neg14, Neg11 and Neg12. The N2-09, N2-10, N2-11
SSWL parameters describe whether the property that the definite NP (noun phrase) contains additional 
markers which is absent in the indefinite NP (N2-09, expressed in Niger-Congo languages like Basaa,
or in the Eastern Armenian language), whether the Numeral has a different form in definite and 
indefinite contexts (N2-10, expressed in Arabic and Hebrew, in Icelandic and in Arawakan languages 
like Garifuna), and whether the noun itself has a different form in definite and 
indefinite contexts (N2-11, expressed for instance in Eastern Armenian, Danish, Icelandic, 
Norwegian, and in the Sandawe language). The two-vertex component connecting N2-01 and N2-03
also pertain to the same subset of SSWL variables: N2-01 is expressed if 
at least one numeral can precede the noun in an indefinite NP, while N2-03 is expressed
if the same occurs with definite NP. 

\smallskip

The parameter Neg13, Distinct Negation of Existence, is expressed in a language when 
negation of existence differs from Standard Negation (this is expressed in
many languages including for instance Arabic and Hebrew, Mandarin, 
Hungarian, Japanese), while Neg 14, Distinct Negation of Location, is 
expressed if the negation of predications of location differs from Standard Negation
and it also tends to be expressed in the same languages in which Neg13 is
expressed. The edge connecting Neg11 and Neg12, on the other hand, connect
the Distinct Negation of Class/Property (Neg11, expressed for example in Arabic, 
Burmese, Fijian, Kiswahili) 
where negation of predications of class inclusion and property assignments 
differs from Standard Negation, and Distinct Negation of identity (Neg12, expressed
for example in Arabic and Hebrew, or in Indonesian, Thai, Kiswahili)
where negation used in predications of identity differs from Standard Negation. 
At these $\epsilon$-scales these parameters in the Negation sector of the SSWL data
do not yet coalesce with the Neg07-Neg10 connected component discussed above. 
Another single edge component relating V2-01 and V2-02 connects the
Declarative Verb-Second property (V2-01, expressed for instance in most 
Germanic languages, in Estonian, in the Austroasiatic Khasi language, 
or the Malayo-Polynesian Bajau language) 
which is expressed when a language allows only one constituent to 
precede the finite verb in declarative main clauses and the
Interrogative Verb-Second (V2-02, expressed for instance in the Germanic languages,
in Spanish, Armenian, Georgian, in the Niger-Congo Dagaare language, or 
the Austroasiatic Khasi language)
which is expressed when a language allows only one wh-constituent to 
precede the finite verb in interrogative main clauses. 
The fact that the Germanic subfamily of the Indo-European family
shares both features may be a factor in driving the connection seen 
at this scale in the graph. The remaining two vertex connection 
visible at this $\epsilon$-scale relates the 04 and 06 parameters,
that is, Object Verb, expressed when a verb can follow its object in a neutral context,
and Subject Object Verb (SOV), expressed when the order Subject Object Verb can be used in a neutral context. 
Note that other similar pairs in the word order sector of the SSWL database, such as 01
Subject Verb and 05 Subject Verb Object (SVO) do not yes form connected components
at these scales, while Object Verb and Subject Object Verb already do. This may reflect
the fact that SOV languages are slightly more abundant (around 45$\%$ of world languages)
than SVO languages (around 42$\%$ among world languages).

\smallskip
\subsection{Nearest neighbor structures in syntactic data} 

\begin{figure}[h!]
\begin{center}
\includegraphics[scale=0.95]{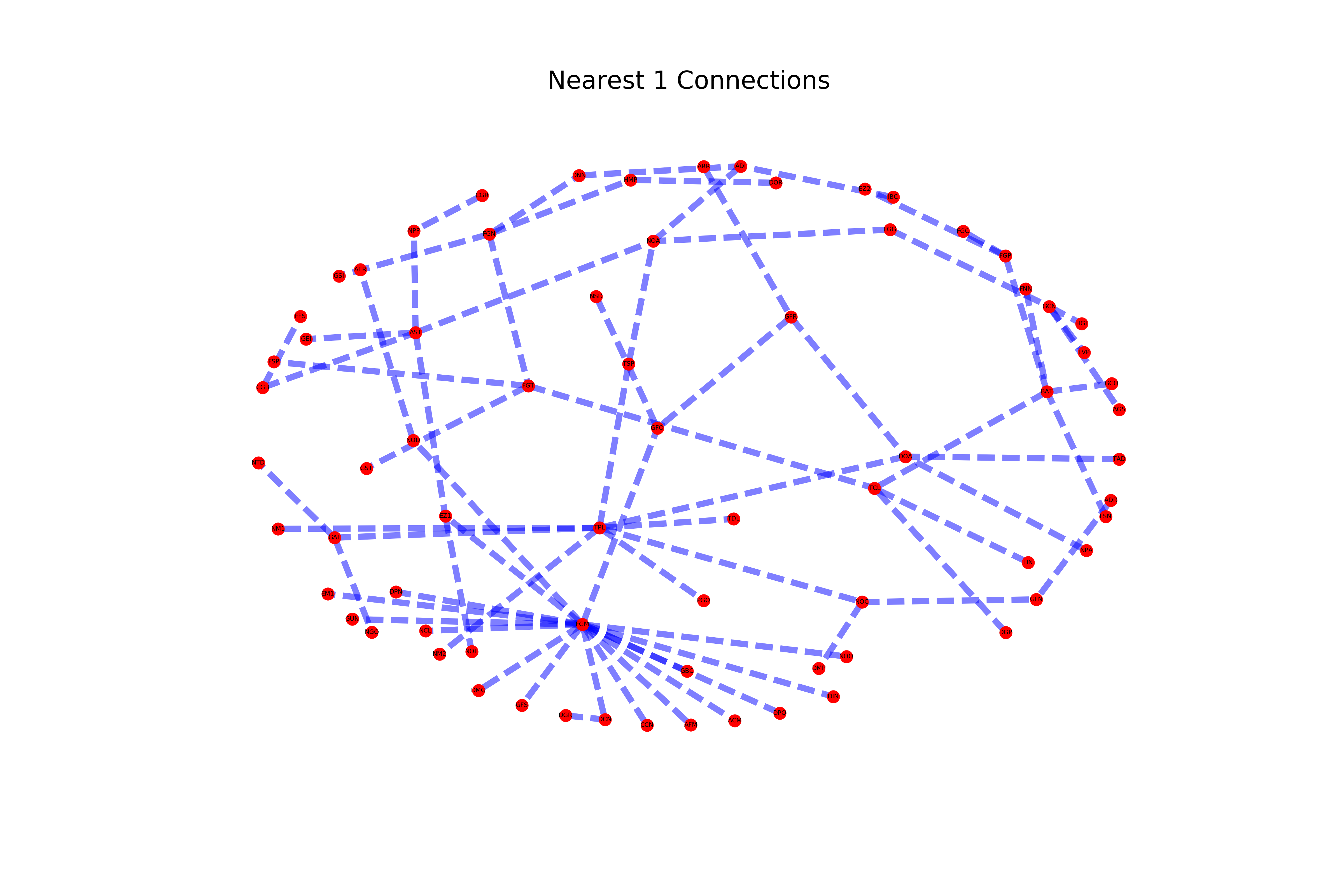}
\includegraphics[scale=0.95]{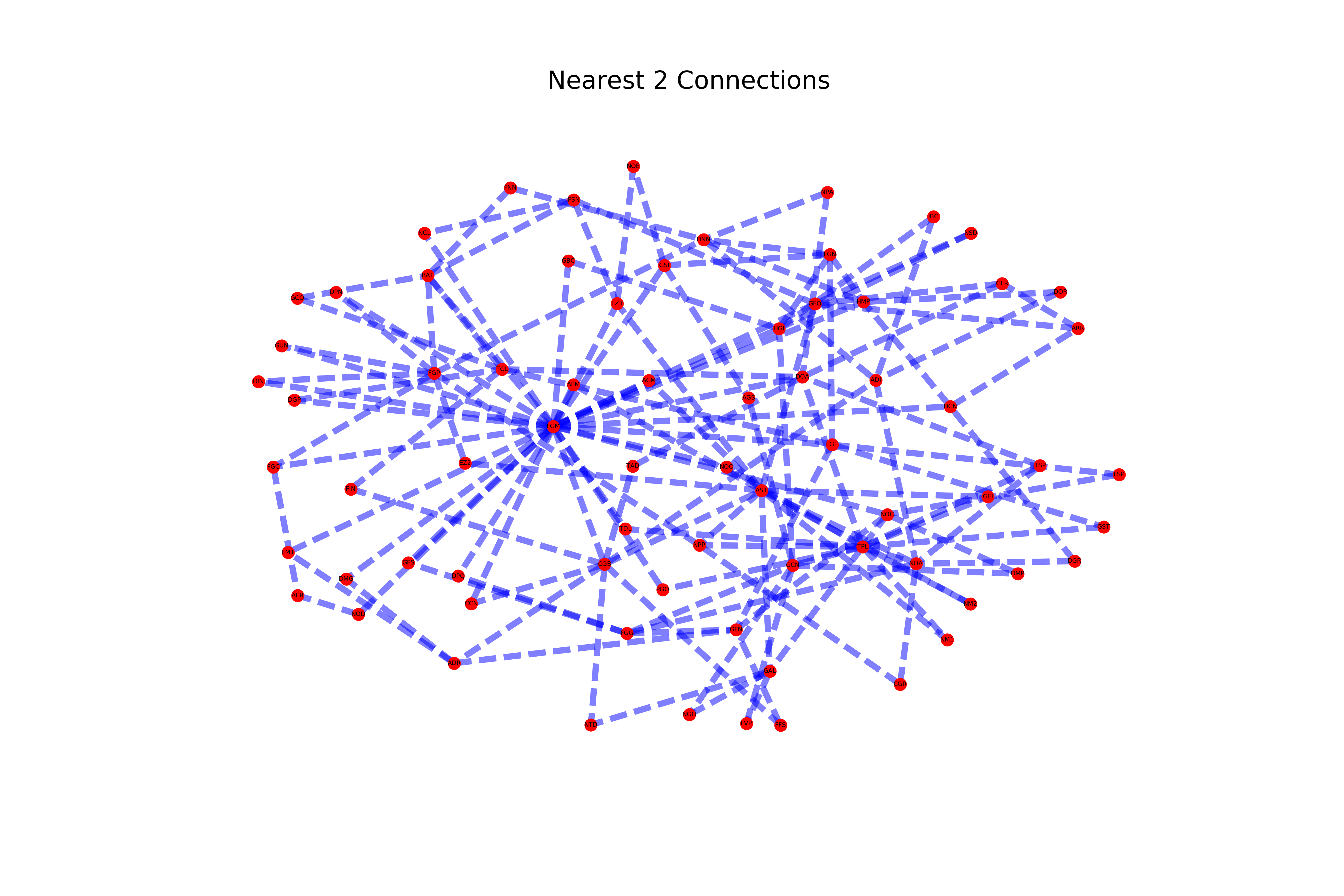}
\caption{$n$-nearest neighbor  graphs for the Longobardi dataset with $n=1$ and $n=2$. \label{L12nFig}}
\end{center}
\end{figure}

\begin{figure}[h!]
\begin{center}
\includegraphics[scale=0.95]{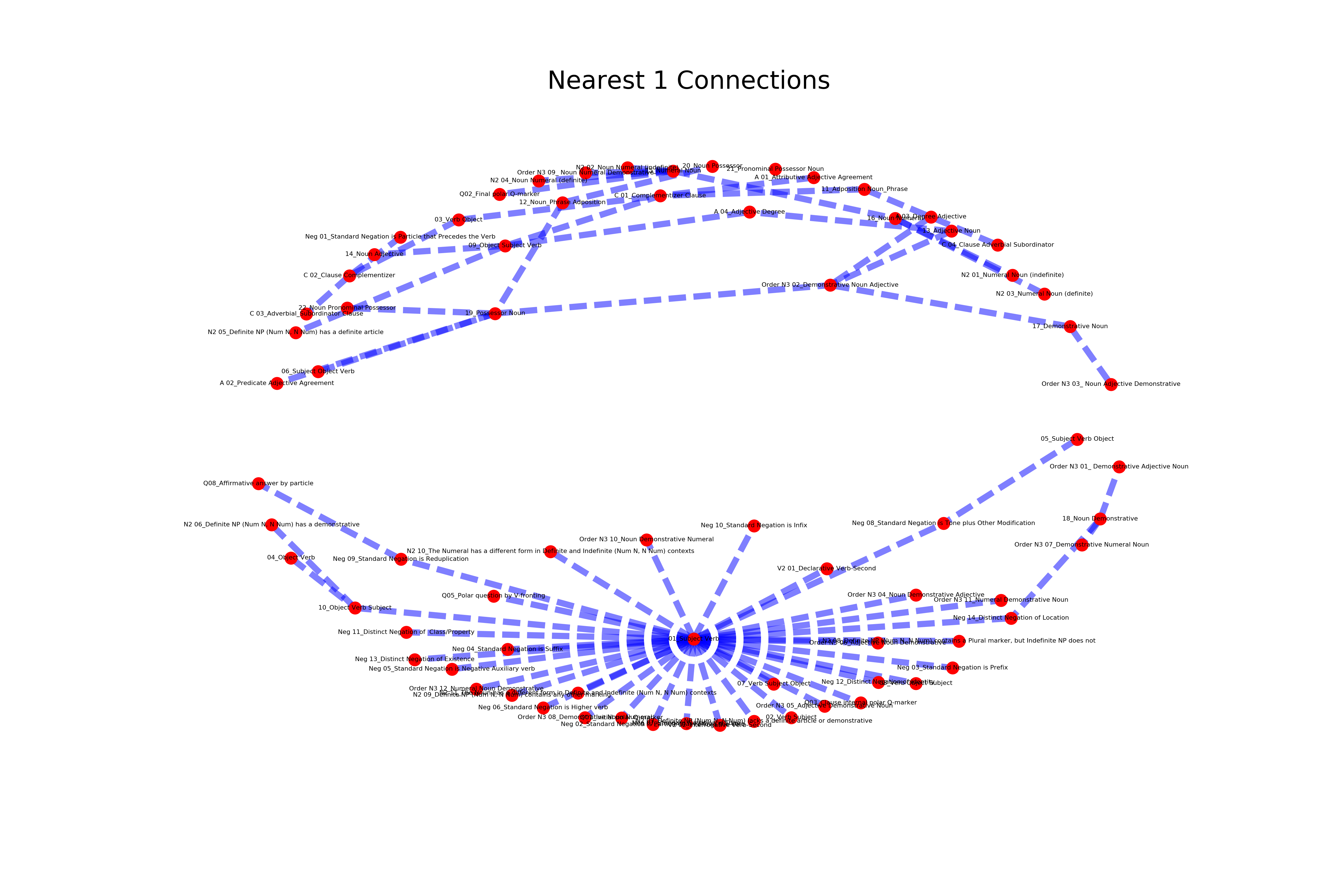}
\includegraphics[scale=0.95]{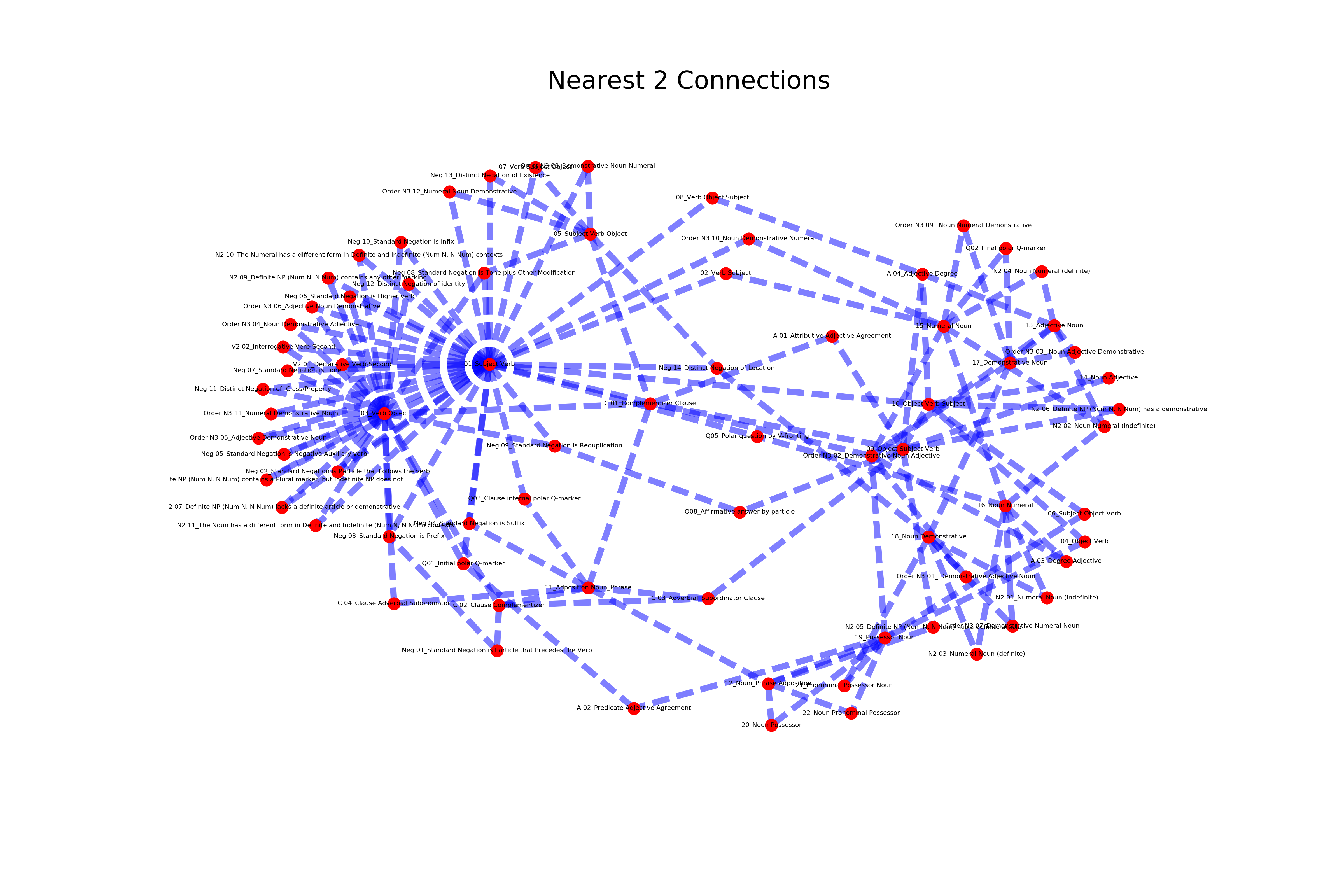}
\caption{$n$-nearest neighbor  graphs for the SSWL dataset with $n=1$ and $n=2$. \label{SSWL12nFig}}
\end{center}
\end{figure}

As we discuss more in detail in \S \ref{ConnClustSec} below,
if we use the $n$-nearest neighbor construction of graphs in
the Belkin--Niyogi algorithm instead of the $\epsilon$-neighborhood
method, the two sets of data, SSWL and Longobardi's, tend to behave
more similarly. 

\smallskip

The nearest one and nearest two connections for the Longobardi
dataset are shown in Figure~\ref{L12nFig}. It stands out very clearly
that the FGM node (gramm.~Case) has a high centrality already at the $n=1$ level,
with nodes like AST (structured APs), CGB (unbounded sg N), FGP (gramm.~person) 
and others like TPL, also having high centrality in
the network at the $n=2$ level. 

\smallskip

A similar analysis of the SSWL data is shown in Figure~\ref{SSWL12nFig}. 
The type of connectivity structures one sees with this method differ
significantly from those obtained by the $\epsilon$-neighborhood 
approach, discussed above. At the $n=1$ stage, the SSWL data
separate neatly into two different connected components. The
component shown in the bottom part of the first plot in Figure~\ref{SSWL12nFig}
has a single node of very high centrality, consisting of the 01 Subject Verb parameter, 
and the rest of the component shows very little interconnectivity.
The only nodes in this component that are not 
directly connected to the central node are 
Q08 (connected to Neg09), N2-06 and 04 (both connected to 10), 05 (connected
to Neg08), Order-N3-01 and Order-N3-07 (both connected to 18), while all
other nodes are directly connected to the central 01 Subject Verb node and
most of them are valence one.
Most strikingly, this component is a tree (trivial first Betti number). 
This structure may be interpreted as an indicator of an influence of the value 
of the 01 Subject Verb parameter on the parameters located at all the adjacent nodes. 

\smallskip

The other component, shown at the top of the first plot in Figure~\ref{SSWL12nFig}
has a very different structure. It contains no single node of very high centrality and
it has a higher degree of interconenctivity (a nontrivial first Betti number).
The largest valence of nodes in this component is just $5$ (19 Possessor Noun node);
there are a few nodes of degree $4$
(Order-N3-02 Demonstrative Noun Adjectve node, C01 Complementizer Clause node,  
09 Object Subject Verb) and of degree $3$ 
(A03 Degree Adjective, 15 Numeral Noun, 
22 Noun Pronominal Possessor). Note how the subdivision of nodes into
the two connected components in this $n=1$ graph does not appear to
follow any of the natural subdivisions of the SSWL data into different sectors:
for example, word order parameters like Subject Verb or Subject Object Verb
do not belong to the same component, or the Numeral parameters N2, which 
fall partly in one and partly in the other component. 
Negation parameters all belong to the first component (at the bottom of
the figure) but they do not form any interconnected structure, unlike
the complete graphs in the $\epsilon$-neighborhood picture.  

\smallskip

At the $n=2$ level, the two components have already merged. A second high centrality
node, 03 Verb Object, has appeared alongside 01 Subject Verb. A large number of nodes
are directly connected to these two nodes and to no others.

\section{Exploring the $\epsilon$-$t$ space}

In this section we discuss how the global clustering and a measure of
connectivity for the graphs $G(\epsilon,t)$ generated from the datasets
vary with the parameter $\epsilon$. This provides some insight into how 
the graphs evolve within the Belkin--Niyogi process and at what 
$\epsilon$-value the graphs stabilize to a complete graph.

\smallskip
\subsection{Connectivity and clustering}\label{ConnClustSec}

The measure of connectivity that we consider here is {\em vertex-connectivity},
namely the minimum number of nodes that needs to be removed to make the 
graph disconnected. 
Clustering, on the other hand, is defined as the mean number of triangular sub-graphs 
that can be generated from the neighborhood of any given node.

\begin{figure}[h!]
\begin{center}
\includegraphics[scale=0.65]{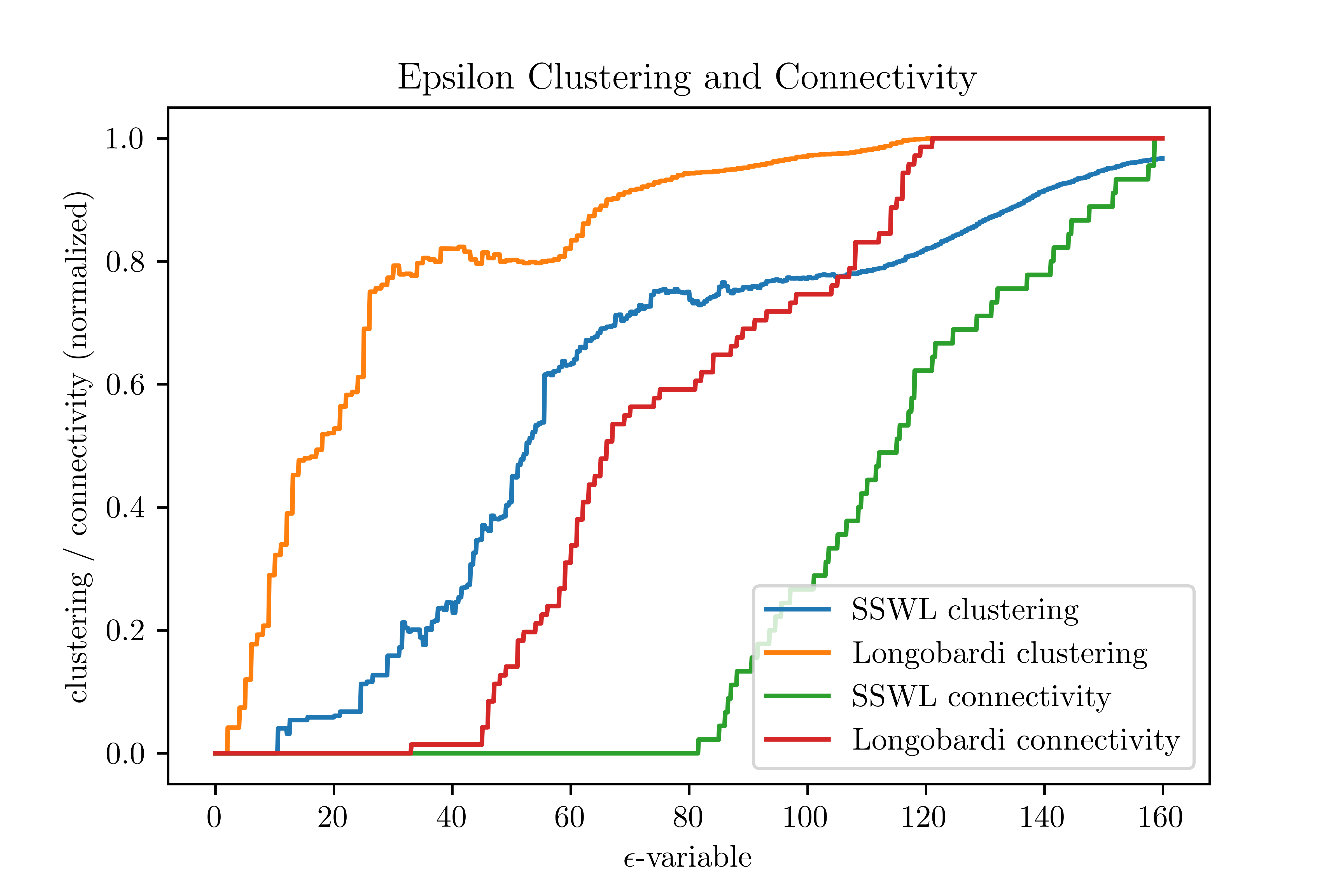}
\includegraphics[scale=0.65]{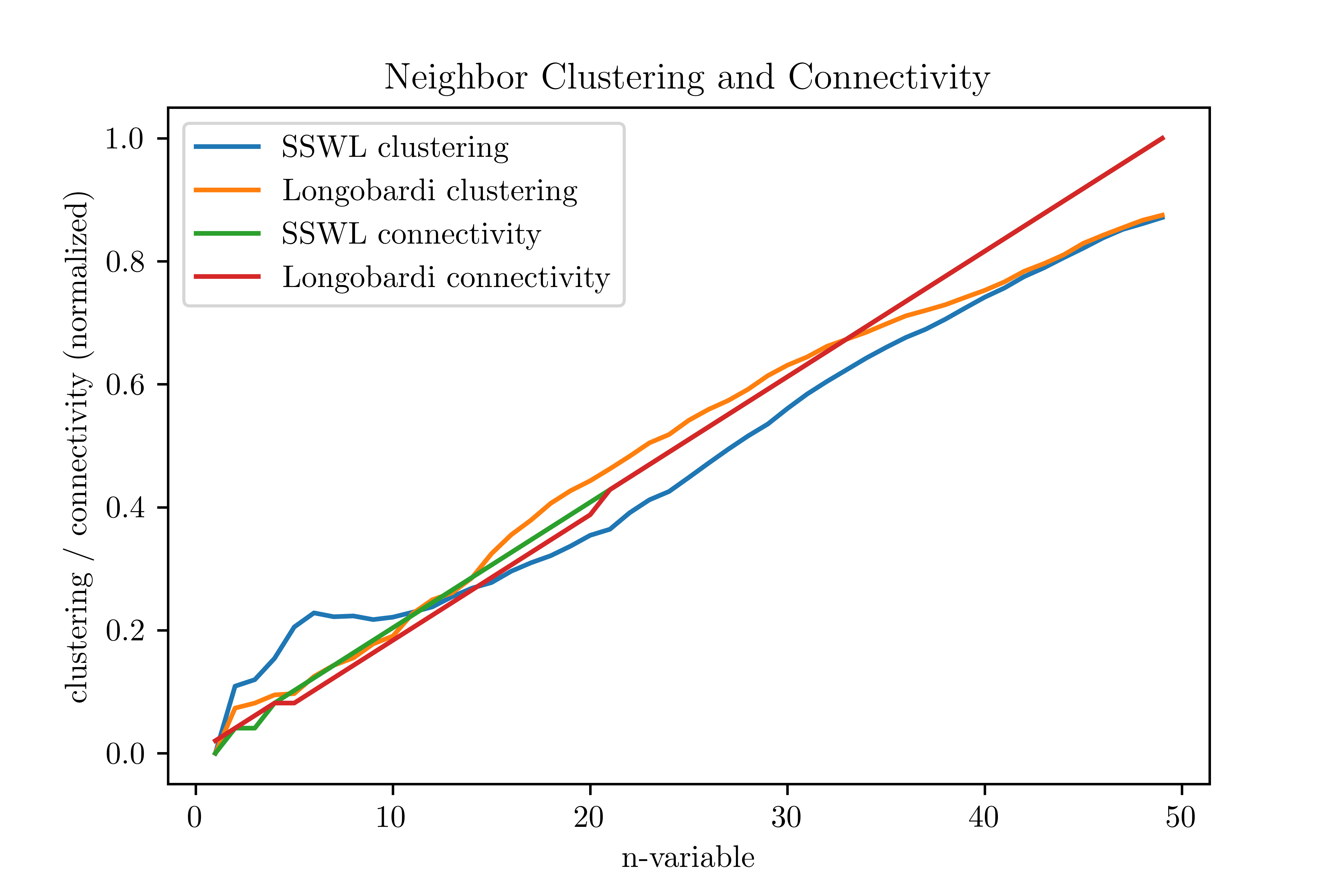}
\caption{Plots of the connectivity and clustering of graphs as a function of 
the parameters $\epsilon$ and $n$, respectively, used in the graph generation methods. \label{CCFig}}
\end{center}
\end{figure}

\smallskip

The behavior of clustering and connectivity, as a function of either the $\epsilon$-variable
in the $\epsilon$-neighborhood construction of the graphs, or of the $n$-variable in the
$n$-nearest neighborhood construction shows a significantly different behavior between
the SSWL data and the Longobardi data in the the $\epsilon$-neighborhood case, 
with the SSWL data exhibiting much lower connectivity and clustering than the Longobardi data.
On the other hand, the behavior of the two datasets appears much more similar in the 
$n$-nearest neighborhood case, see Figure~\ref{CCFig}. If a higher degree of connectivity
and clustering is to be taken as an indicator of the presence of relations between the 
parameters, then the different behavior of the two datasets un the $\epsilon$-neighborhood
construction would point to more structured relations in the Longobardi dataset than in the
set of syntactic variables reported in the SSWL. As we mentioned above (see \cite{Longo1},
\cite{Longo2} for a more detailed explanation), the Longobardi dataset does explicitly record
certain types of entailment relations between the listed parameters, so we know a priori that
the dataset does not consist of independent variables. Previous analysis, such as \cite{Park},
also indicated the presence of relations between the SSWL variables, although relations in the SSWL data are
not explicitly formulated as the relations recorded in the Longobardi data, and were only detected
through a measure of recoverability in Kanerva networks. It is possible that the higher levels of
connectivity and clustering visible in the Longobardi data with the $\epsilon$-neighborhood
method may reflect the more structured type of relations present in the Longobardi data.
It is interesting, though, that when the graph construction is performed using the
$n$-nearest neighborhood method, the two datasets tend to behave much more
similarly, with the SSWL clustering value peaking above the Longobardi value for small $n$
and trailing slightly below for larger values of $n$. 

\smallskip
\subsection{Activity regions in the $\epsilon$-$t$ space}

We investigate here the simultaneous dependence on both the $\epsilon$-variable of the 
$\epsilon$-neighborhood construction of the graphs and the $t$-variable of the heat kernel. 

\begin{figure}[h!]
\begin{center}
\includegraphics[scale=0.65]{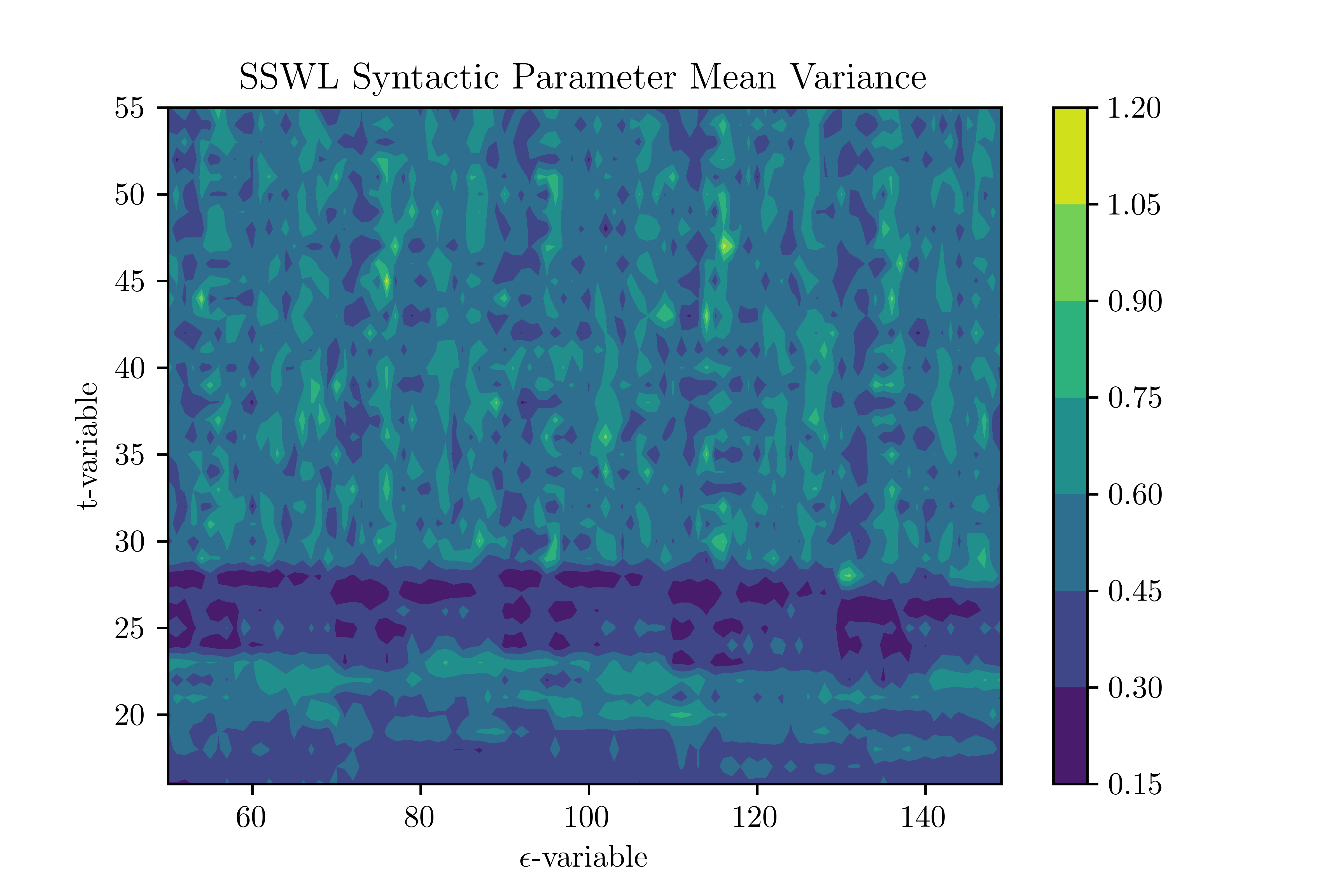}
\includegraphics[scale=0.65]{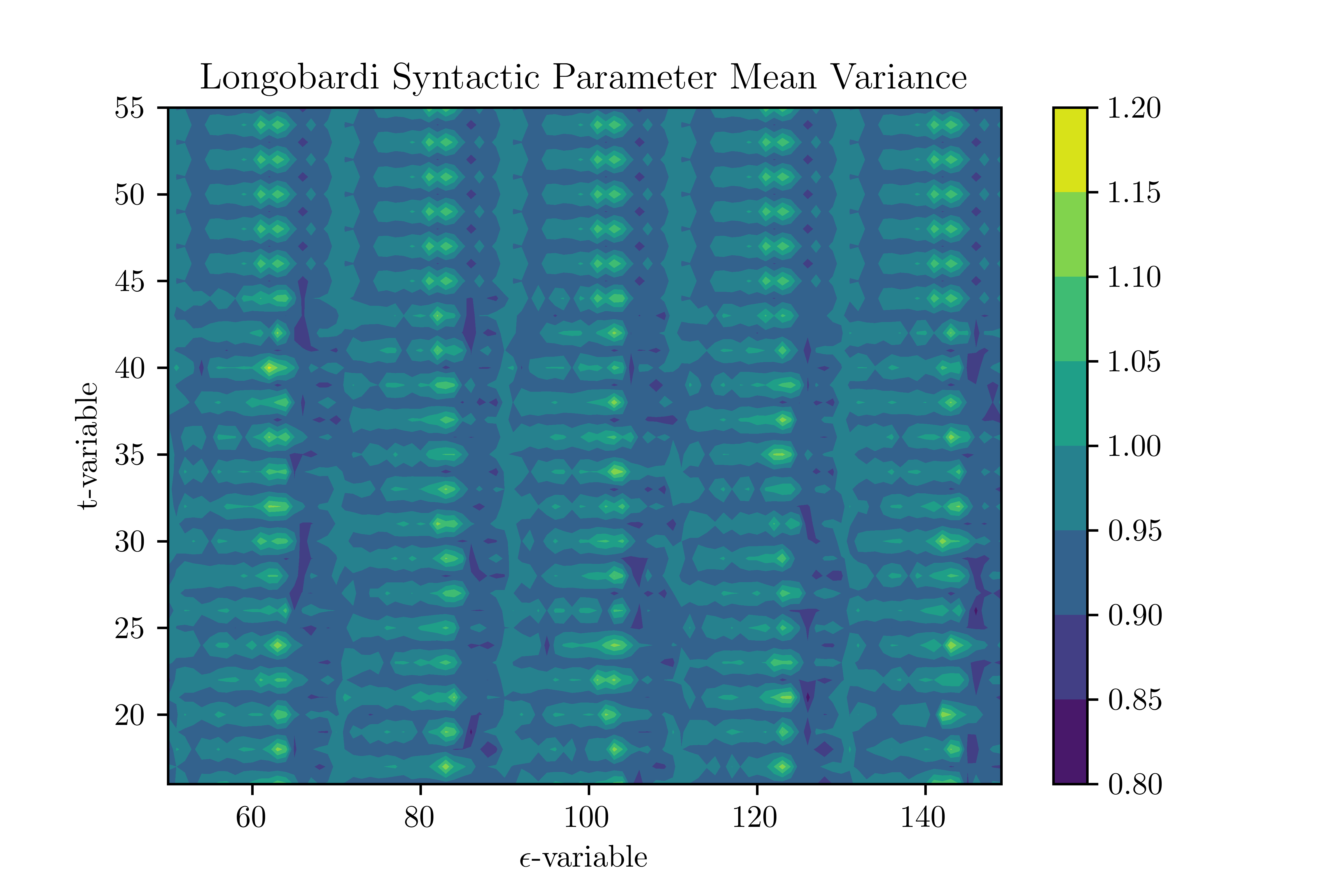}
\caption{Contour plots of the variance of syntactic parameters as a function of the graph parameters $\epsilon$ and $t$. The yellow points indicate peaks in variance of the syntactic parameters. \label{ContourFig}}
\end{center}
\end{figure}

\smallskip

The $\epsilon$-$t$ parameter space was explored to determine which values of $\epsilon$ 
and $t$ give rise to {\em high variance} in the distribution of each parameter under the mapping 
determined by the linear transformation $T$ of \eqref{mapLaplace}, \eqref{Tmap} associated
to a given weighted graph $G(\epsilon,t)$. The reason for considering this high variance 
condition is similar to the usual argument in the setting of principal component analysis, where
high variance is used as an indicator that the resulting variables are highly independent.
Thus, the high variance regions we identify in the $\epsilon$-$t$ space should be regarded
as choices of the $\epsilon$-$t$ parameters that optimize the Belkin--Niyogi representation
in the sense that the Laplace eigenfunction projections provide a set of highly independent variables
for the representation of syntactic parameters. The resulting contour plot identifying high variance
regions is shown in Figure~\ref{ContourFig}.

\begin{figure}[!ht]
\begin{center}
\includegraphics[scale=0.65]{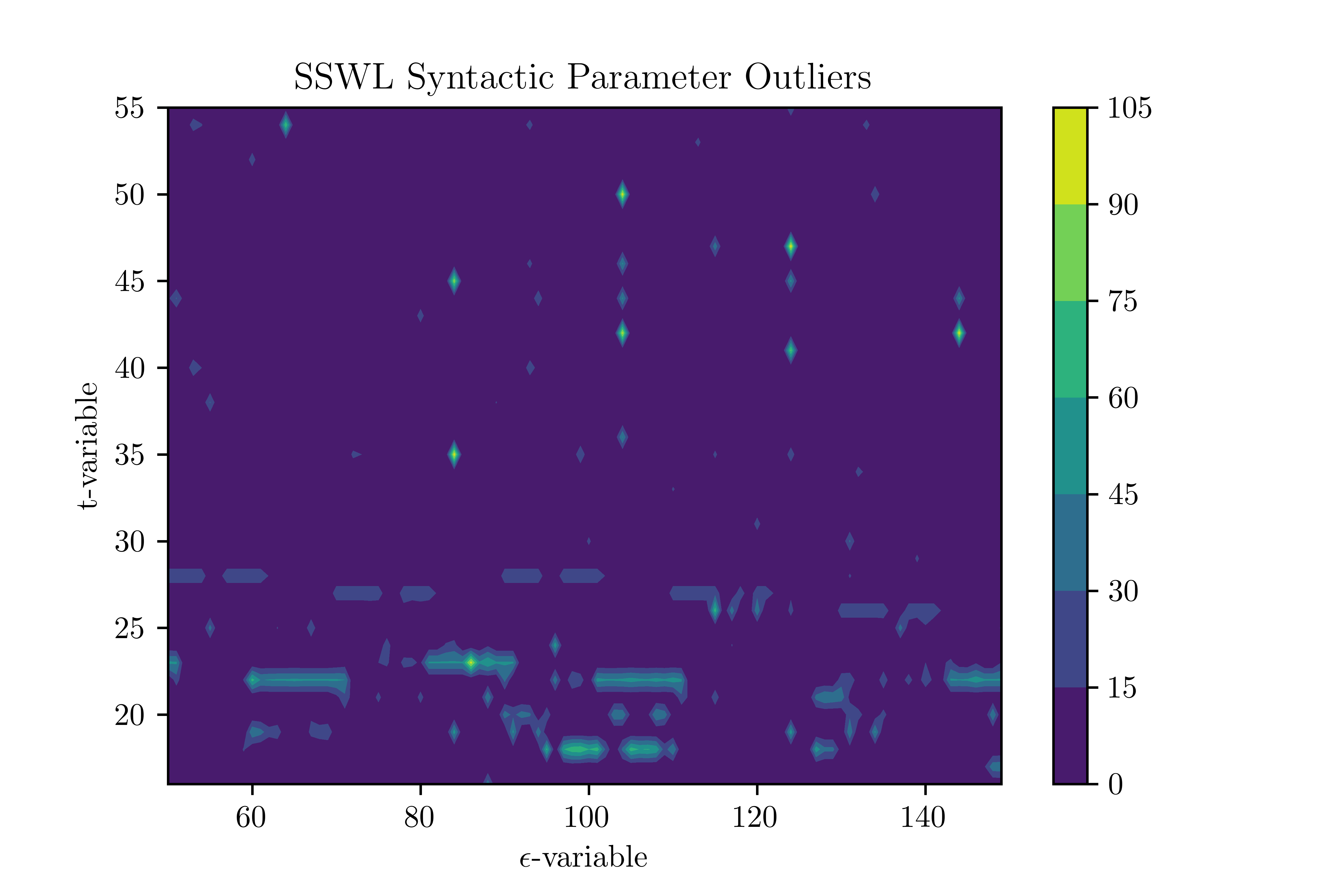}
\includegraphics[scale=0.65]{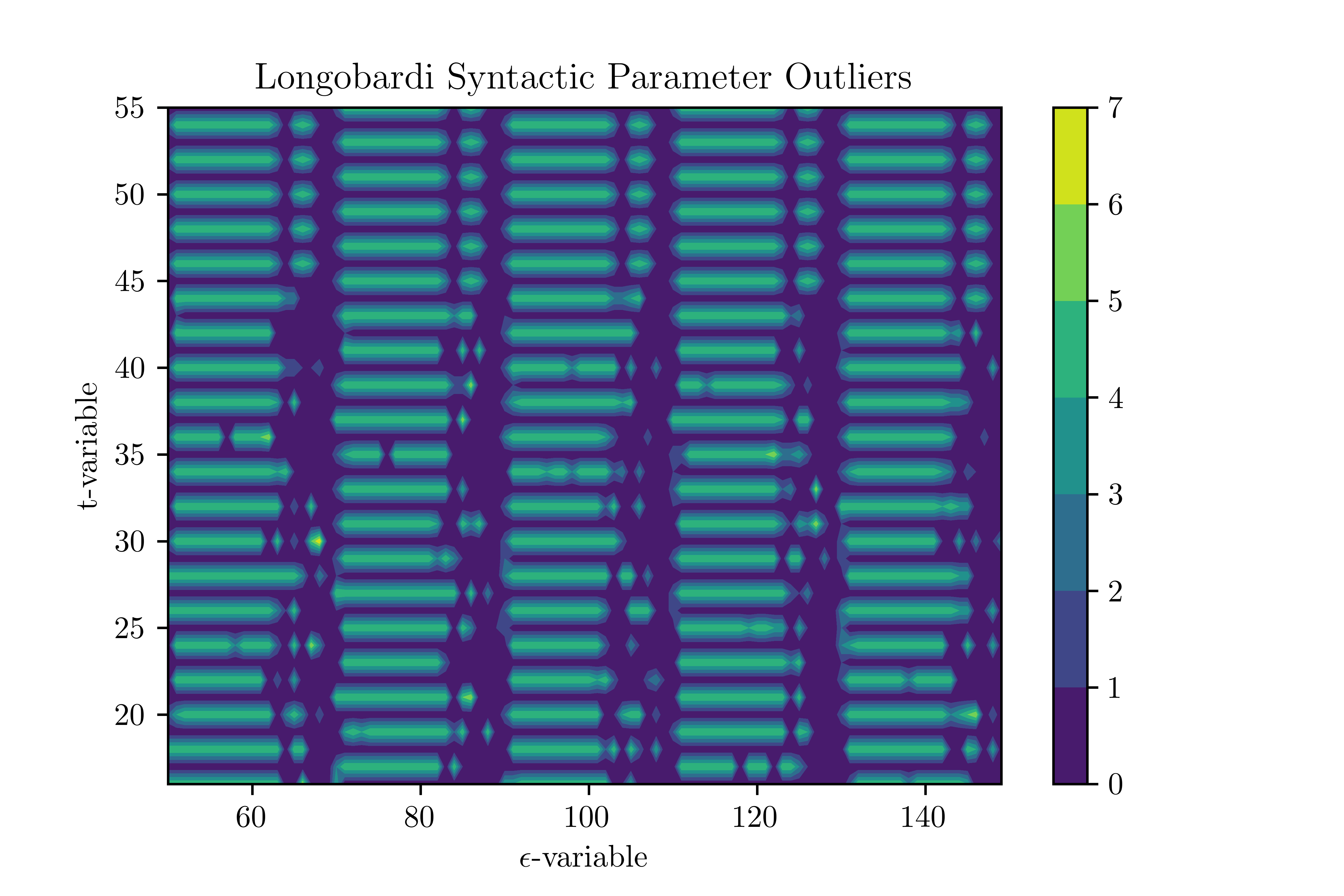}

\caption{Contour plots of the number of outliers of syntactic parameters 
as a function of the graph parameters $\epsilon$ and $t$. \label{OutlierFig}}
\end{center}
\end{figure}

\smallskip

Another test aimed at identifying especially interesting regions in the $\epsilon$-$t$ space
was conducted using a measure of the number of outliers produced among the set of coordinates 
for a given parameter. This measure indicated similar high magnitude in the regions analyzed with
the previous method.

\smallskip

Let $\cL$ be the set of languages in the database under consideration. We regard each $\ell_i \in \cL$
as a binary vector $\ell_i \in \bF_2^n$, where $n$ is the total number of syntactic parameters in the database,
$\ell_i =( p_{i,j} )_{j=1}^n$, where $p_{i,j}$ is the value that the $j$-th parameter takes in the $i$-th language.
Similarly, we view each parameter $p_j$ as a vector $p_j \in \bF_2^N$, where $N=\# \cL$ is the number
of languages in the database, $p_j =(p_{i,j})_{i=1}^N$. 

\smallskip

Counting the number of outliers of the set $p_j =\{ p_{i,j}\, | \, \ell_i \in \cL \}$ and averaging that measure 
over all the sets of parameters gives a new measure of how the Laplacian eigenfunctions map method 
for dimensional reduction improves the efficiency of the new parameters to describe the languages in
$\cL$. The resulting plot of the number of outliers is given in Figure~\ref{OutlierFig}. 

\smallskip
\subsection{Clustering coefficients}

We compute clustering coefficients of the nodes of the graphs $G(\epsilon,t)$ 
using the NetworkX package \cite{NetworkX}.

\smallskip

Consider an undirected graph $G_i = (V,E)$ composed of its set of vertices $V=V(G)$ (nodes) and set of edges
$E=E(G)$, generated from a chosen region of the $\epsilon$-$t$ parameter space identified according
to the analysis described in the previous subsections. The neighborhood, $\mathbb{V}_i$ of any given node, $v_i \in V(G)$ is the set of all nodes connected to that node. The size of $\mathbb{V}_i$ is referred to as the degree (valence) of the vertex $d_i=\deg(v_i)=\# \mathbb{V}_i$. Since we are dealing with graphs that do not have parallel edges (are not multigraphs), the valence also counts the number of edges connected to $v_i$. 
The cardinality 
\begin{equation}\label{Ki}
K_i =\# \{ e\in E(G)\,|\, \partial(e)\cap \mathbb{V}_i \neq \emptyset \}
\end{equation}
of the set all the edges connected to any of the vertices $v_k \in \mathbb{V}_i$ gives a measure of the 
{\em clustering} in the neighborhood of the vertex $v_i$. 
The local clustering coefficient $C_i$ of a vertex $v_i$ is then given by 
\begin{equation}\label{clusterCi}
 C_i = \frac{K_i }{{d_i \choose 2}}
\end{equation} 
Using the NetworkX package, the values $K_i$ can be determined from the graph objects present 
in the environment and the degrees $d_i$ are immediately available for any given node. 

\smallskip

\begin{figure}[h!]
\begin{center}
\includegraphics[scale=0.65]{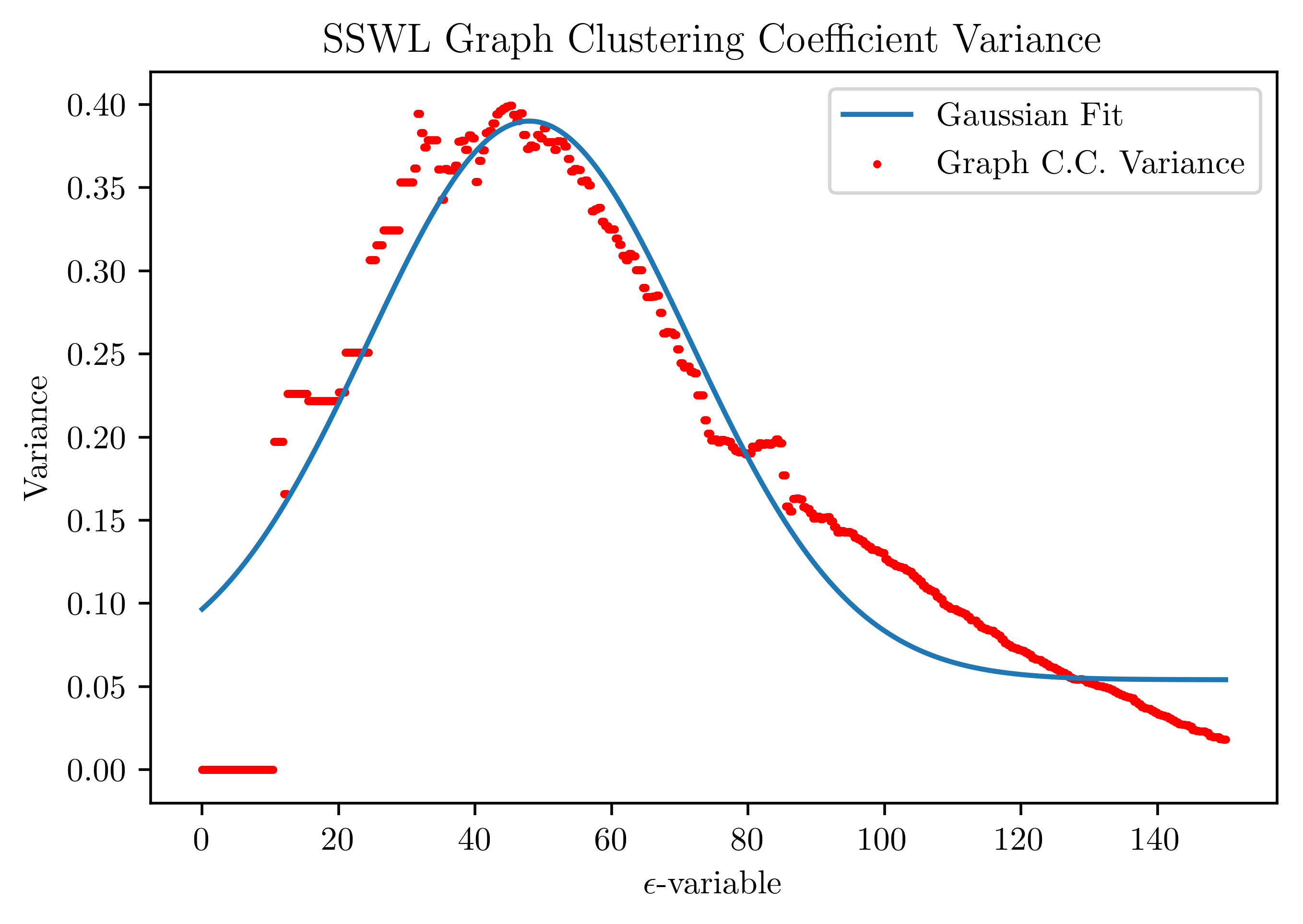}
\includegraphics[scale=0.65]{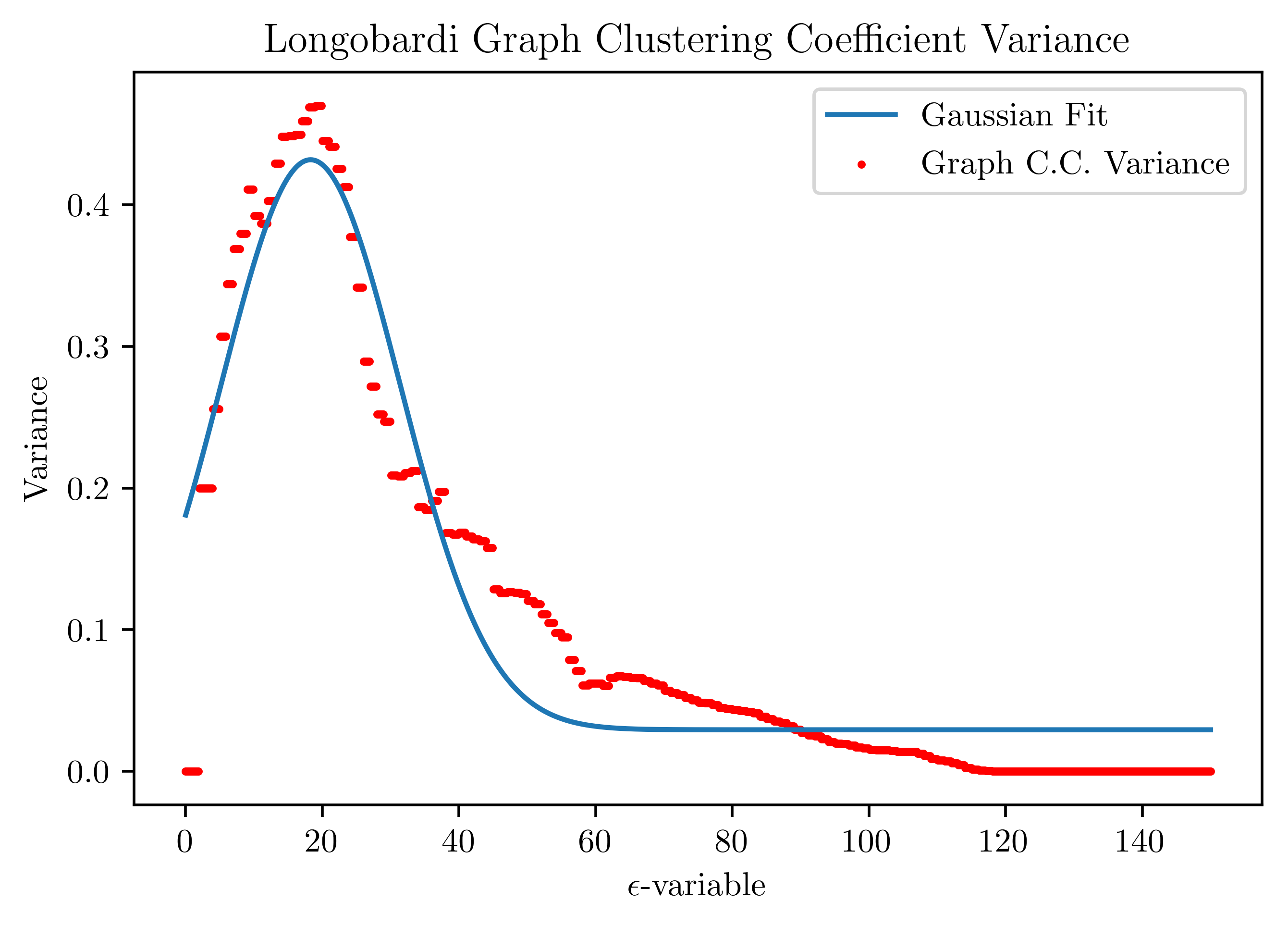}
\caption{A plot of the variance among the clustering coefficents of a graph $G(\epsilon)$ as a 
function of $\epsilon$. \label{GaussFig} }
\end{center}
\end{figure}

The variance of the clustering coefficients $C_i$ as a function of the $\epsilon$-variable in
the $\epsilon$-neighborhood construction of the graph are well approximated by a 
Gaussian law of the form
$$ f(x) = A \exp\left({-\frac{(x-H)^2}{\sigma^2}}\right) + V, $$
with the parameters as indicated in the Table, see Figure~\ref{GaussFig}.  
\begin{center}
\begin{table}[h!]
\begin{tabular}{ |c|c|c| } 
 \hline
 Fit Parameter & Value & 1$\sigma$ Error ($\%$) \\ 
 \hline
 $A$ & 0.3359 & 0.3359  \\ 
  \hline
 $H$ & 47.9530 & 22.99  \\ 
 \hline
 $\sigma$ & 33.35 & 43.78  \\ 
 \hline
 $V$ & 0.0541 & 0.002310  \\ 
\hline
\end{tabular}
\smallskip
\caption{SSWL Gaussian-Fit Parameter Data}
\end{table}
\end{center}

\begin{center}
\begin{table}[h!]
\begin{tabular}{ |c|c|c| } 
 \hline
 Fit Parameter & Value & 1$\sigma$ Error ($\%$) \\ 
 \hline
 $A$ & 0.4025 & 0.4070  \\ 
  \hline
 $H$ & 18.2832 & 15.74  \\ 
 \hline
 $\sigma$ & 18.5040 & 26.10  \\ 
 \hline
 $V$ & 0.02934 & 0.1496  \\ 
\hline
\end{tabular}
\smallskip
\caption{Longobardi Gaussian-Fit Parameter Data}
\end{table}
\end{center}

\begin{figure}[h!]
\begin{center}
\includegraphics[scale=0.65]{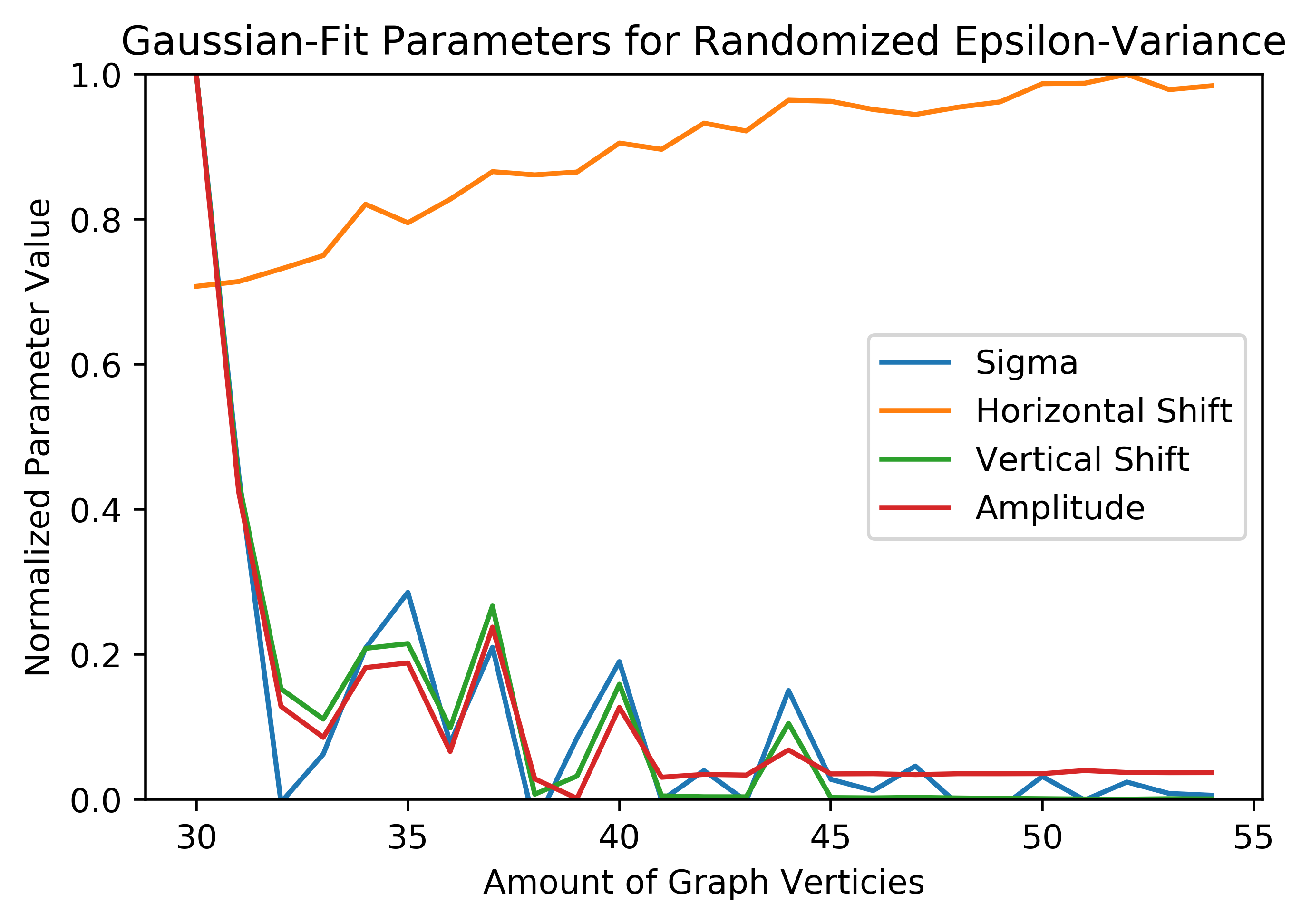} 
\caption{Gaussian parameters $\sigma$, $H$, $V$, and $A$ for random 
graphs as a function of the number of vertices. \label{ParamPlot}}
\end{center}
\end{figure}

\smallskip

We compare this behavior with that of random graphs with a varying number of vertices,
obtained by generating random binary $n\times 260$ matrices where $n$ is the number 
of graphs vertices and using them as parameter data to generate graphs.
A plot of the values of the Gaussian parameters $\sigma$, $H$, $V$, and $A$
as a function of the number of vertices in the graph is shown in 
Figure~\ref{ParamPlot}. 
The convergence of the horizontal shift $H$ indicates that this value only partially depends 
on the number of vertices in the graph and possibly on the number of parameter values at 
each vertex. The values of $\sigma$ and of the amplitude $A$ appear to converge to small 
but non-zero values, and the vertical shift value $V$ appears to converge to zero. 

\smallskip

The set of all clustering coefficients $C_i$ can be used further in the context of phylogenetic reconstructions
of language families, as outlined in the last section of \cite{OSBM}. Namely, the phylogenetic models 
used in \cite{OSBM} based on the method of phylogenetic algebraic geometry, are 
based on considering individual parameters as identically distributed independent random variables
evolving according to the same Markov process on a tree. This hypothesis is not appropriate in view
of the many relations that exist between syntactic parameters (some known explicitly, as those described
in \cite{Longo1}, \cite{Longo2}) and others only detected statistically, as discussed in the present paper,
or with a different method in \cite{Park}. A modification of the phylogenetic model can be obtained
by introducing different weights attached to different parameters in the boundary distribution at the
leaves of the tree, which gives more weight in the model to parameters that are more likely to be
``independent variables" and less weight to those that have a high degree of dependency on others.
The latter can be measured in different possible ways, for instance through a degree of recoverability
in Kanerva network as was done in \cite{Park}, or through a function of the clustering coefficients
computed here. We will return to investigate this kind of application in future work.

\end{document}